\documentclass[11pt,a4paper]{article}
\usepackage[hyperref]{acl2021}
\usepackage{longtable,booktabs}
\usepackage{times}
\usepackage{amsmath,amsfonts,bm}

\def\eqref#1{equation~\ref{#1}}
\def\1{\bm{1}}

\def\mD{{\bm{D}}}

\def\mP{{\bm{P}}}

\DeclareMathAlphabet{\mathsfit}{\encodingdefault}{\sfdefault}{m}{sl}
\SetMathAlphabet{\mathsfit}{bold}{\encodingdefault}{\sfdefault}{bx}{n}

\newcommand{\R}{\mathbb{R}}

\DeclareMathOperator*{\argmax}{arg\,max}
\DeclareMathOperator*{\argmin}{arg\,min}

\usepackage{hyperref}
\usepackage{url}
\usepackage{multirow}
\usepackage[ruled,vlined]{algorithm2e}
\usepackage{amsmath}
\usepackage{amsfonts}
\usepackage{amssymb}
\usepackage{amsthm}
\def\aclpaperid{3648}
\aclfinalcopy
\usepackage{setspace}
\usepackage{graphicx}
\usepackage{xcolor}
\usepackage{xspace}
\usepackage[textwidth=1.2in]{todonotes}
\usepackage{dsfont}
\newcommand*{\email}[1]{\texttt{#1}}
\newcommand{\method}{VOLT\xspace}

\title{Vocabulary Learning via Optimal Transport  \\ for Neural Machine Translation  }  

\author{
Jingjing Xu$^{1}$, \ 
Hao Zhou$^{1}$, \ 
Chun Gan$^{1,2}$\footnotemark[2], \ 
Zaixiang Zheng$^{1,3}$\footnotemark[2], \
Lei Li$^{1}$ \
\\ 
$^{1}$ByteDance AI Lab \\
$^{2}$Math Department, University of Wisconsin–Madison \\
$^{3}$Nanjing University \\
 \email{jingjingxu@pku.edu.cn } \\
 \email{zhouhao.nlp@bytedance.com } \\
 \email{cgan5@wisc.edu } \\
 \email{zhengzx@smail.nju.edu.cn } \\
  \email{lilei@ucsb.edu} \\
}

\begin{document}

\maketitle

\renewcommand{\thefootnote}{\fnsymbol{footnote}}

\begin{abstract}
The choice of token vocabulary affects the performance of machine translation.   
This paper aims to figure out what is a good vocabulary and whether one can find the optimal vocabulary without trial training. To answer these questions, we first provide an alternative understanding of the role of vocabulary from the perspective of information theory. Motivated by this, we formulate the quest of vocabularization -- finding the best token dictionary with a proper size -- as an optimal transport (OT) problem.  We propose
\textbf{VOLT}, a simple and efficient solution without trial training. 
Empirical results show that \method  outperforms  widely-used vocabularies  in  diverse  scenarios, including WMT-14 English-German and TED multilingual translation. For example, \method achieves almost 70\% vocabulary size reduction and 0.5 BLEU gain on English-German translation. Also,  compared to BPE-search, \method reduces the search time from 384 GPU hours to 30 GPU hours on English-German translation. Codes are available at \url{https://github.com/Jingjing-NLP/VOLT}.

\end{abstract}
\footnotetext[2]{This work is done during the internship at ByteDance AI Lab.}

\section{Introduction}
\label{sec:intro}

% intro from zhouh

Due to the discreteness of text,  vocabulary construction~( vocabularization for short) is a prerequisite for neural machine translation (NMT) and many other natural language processing~(NLP) tasks using neural networks~\citep{DBLP:conf/nips/MikolovSCCD13,attention, DBLP:conf/emnlp/GehrmannDR18,DBLP:journals/widm/ZhangWL18,DBLP:conf/naacl/DevlinCLT19}.  
Currently, sub-word approaches like Byte-Pair Encoding (BPE) are widely used in the community~\citep{scalingnmt,DBLP:conf/mtsummit/DingRD19,deeptransformer}, and achieve quite promising results in practice~\citep{DBLP:conf/acl/SennrichHB16a, DBLP:conf/acl/Costa-JussaF16,DBLP:journals/tacl/LeeCH17,DBLP:conf/emnlp/KudoR18, DBLP:conf/aaai/Al-RfouCCGJ19, DBLP:conf/aaai/WangCG20}. The key idea of these approaches is selecting the most frequent sub-words~(or word pieces with higher probabilities) as the vocabulary tokens.
In information theory, these frequency-based approaches are simple forms of data compression to reduce entropy~\citep{gage1994new}, which makes the resulting corpus easy to learn and predict~\citep{martin2011mathematical,bentz2016word}.

%current approaches only consider frequency (or entropy) as the main criteria while current approaches only consider frequency (or entropy) as the main criteria, while 
However, the effects of vocabulary size are not sufficiently taken into account since current approaches only consider frequency (or entropy) as the main criteria.
Many previous studies~\citep{sennrich-zhang-2019-revisiting, DBLP:conf/mtsummit/DingRD19,DBLP:conf/acl/ProvilkovEV20, DBLP:journals/mt/SaleskyRCNN20} show that vocabulary size also affects downstream performances, especially on low-resource tasks. Due to the lack of appropriate inductive bias about size, trial training (namely traversing all possible sizes) is usually required to search for the optimal size, which takes high computation costs. For convenience,
most existing studies only adopt the widely-used settings in implementation. 
For example, 30K-40K is the most popular size setting in all 42 papers of Conference of Machine Translation (WMT) through 2017 and 2018~\citep{DBLP:conf/mtsummit/DingRD19}. %In this work, we aim to 

In this paper, we propose to explore automatic vocabularization by simultaneously considering entropy and vocabulary size without expensive trial training.
Designing such a vocabularization approach is non-trivial for two main reasons.
First, it is challenging to find an appropriate objective function to optimize them at the same time. Roughly speaking, the corpus entropy decreases with the increase of vocabulary size, which benefits model learning~\citep{martin2011mathematical}. On the other side, too many
tokens cause token sparsity, which hurts model learning~\citep{DBLP:conf/tsd/AllisonGG06}. 
Second, supposing that an appropriate measurement is given, it is still challenging to solve such a discrete optimization problem due to the exponential search space.

 \begin{figure}[t]
     \centering
     \includegraphics[width=0.9\linewidth]{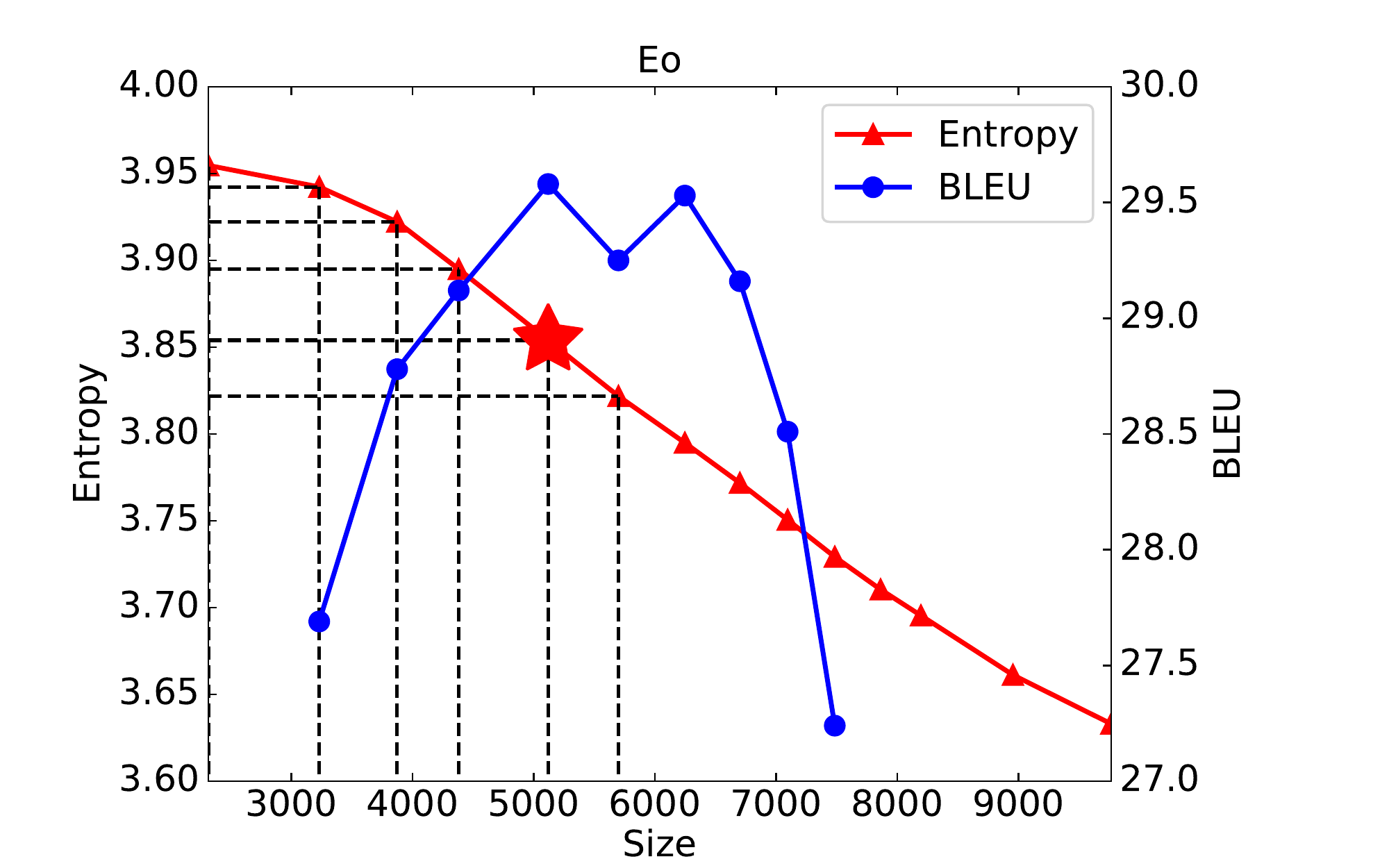}
     \caption{ An illustration of marginal utility. We sample  BPE-generated vocabularies with different sizes from
Eo-En translation and draw their entropy (See Eq.2) and BLEU lines. ``{\textcolor{red}{Star}}'' represents the vocabulary with the maximum marginal utility. Marginal utility (See Eq.1) evaluates the increase of benefit (entropy decrease) from an increase of cost (size).  }
     \label{fig:muv}
 \end{figure}
 
To address the above problems, we propose a \textbf{VO}cabulary \textbf{L}earning approach via optimal \textbf{T}ransport, \method for short. It can give an appropriate vocabulary in polynomial time by considering corpus entropy and vocabulary size.
Specifically, given the above insight of contradiction between entropy and size, we first borrow the concept of \textit{Marginal Utility} in economics~\citep{samuelson1937note} and propose to use \textit{Marginal Utility of Vocabularization}~(MUV) as the measurement.
The insight is quite simple: in economics, marginal utility is used to balance the benefit and the cost and we use MUV to balance the entropy (benefit) and vocabulary size (cost). Higher MUV is expected for Pareto optimality. Formally, MUV is defined as the negative derivative of entropy to vocabulary
size. 
Figure~\ref{fig:muv} gives an example about marginal utility.
Preliminary results verify that MUV correlates with the downstream performances on two-thirds of tasks (See Figure~\ref{fig:mubleu}).

Then our goal turns to maximize MUV in tractable time complexity. 
We reformulate our discrete optimization objective into an optimal transport problem~\citep{DBLP:conf/nips/Cuturi13}  that can be solved in polynomial
time by linear programming.
Intuitively, the vocabularization process can be regarded as finding the \textit{optimal transport matrix} from the\textit{ character distribution} to the \textit{vocabulary token distribution}.
Finally, our proposed \method will yield a vocabulary from the optimal transport matrix.

We evaluate our approach on multiple machine translation tasks, including WMT-14 English-German translation, TED bilingual translation, and TED multilingual translation. Empirical results show that \method beats widely-used vocabularies in diverse scenarios. Furthermore, \method is a lightweight solution and does not require expensive computation resources. On English-German translation,  \method only takes 30 GPU hours to find vocabularies, while the traditional BPE-Search solution takes 384 GPU hours.

\section{Related Work}
\label{sec:related}

Initially, most neural models were built upon word-level vocabularies~\citep{DBLP:conf/acl/Costa-JussaF16,attention,DBLP:conf/ijcai/ZhaoSY19}. While achieving promising results, it is a common constraint that word-level vocabularies fail on handling rare words under limited vocabulary sizes.

%\citet{DBLP:conf/acl/Costa-JussaF16} propose a character-level vocabulary that adopts single characters as the minimum semantic unit. 

Researchers recently have proposed several advanced vocabularization approaches, like byte-level approaches~\citep{DBLP:conf/aaai/WangCG20}, character-level approaches~\citep{DBLP:conf/acl/Costa-JussaF16,DBLP:journals/tacl/LeeCH17,DBLP:conf/aaai/Al-RfouCCGJ19}, and sub-word approaches~\citep{DBLP:conf/acl/SennrichHB16a,DBLP:conf/emnlp/KudoR18}.  
 Byte-Pair Encoding (BPE)~\citep{DBLP:conf/acl/SennrichHB16a} is proposed to get subword-level vocabularies. The general idea is to merge pairs of frequent character sequences to create sub-word units. Sub-word vocabularies can be regarded as a trade-off between character-level vocabularies and word-level vocabularies. Compared to word-level vocabularies, it can decrease the sparsity of tokens and increase the shared features between similar words, which probably have similar semantic meanings, like ``happy'' and ``happier''. Compared to character-level vocabularies, it has shorter sentence lengths without rare words.  Following BPE, some variants recently have been proposed, like BPE-dropout~\citep{DBLP:conf/acl/ProvilkovEV20}, SentencePiece~\citep{DBLP:conf/emnlp/KudoR18}, and so on. 

Despite promising results, most existing sub-word approaches only consider frequency while the effects of vocabulary size is neglected. Thus, trial training is required to find the optimal size, which brings high computation costs. More recently, some studies notice this problem and propose some practical solutions~\citep{DBLP:journals/corr/abs-1810-01480, DBLP:conf/emnlp/CherryFBFM18, DBLP:conf/naacl/Chen0SCYW19, DBLP:journals/mt/SaleskyRCNN20}.

\section{Marginal Utility of Vocabularization} %New finding: RBV Indicates Model Performance
\label{sec:finding}

In this section, we propose to find a good vocabulary measurement by considering entropy and size. As introduced in Section~\ref{sec:intro},
it is non-trivial to find an appropriate objective function to optimize them simultaneously. On one side, with the increase of vocabulary size, the corpus entropy is decreased, which benefits model learning~\citep{bentz2016word}.
On the other side, a large vocabulary causes parameter explosion and token sparsity problems, which hurts model learning~\citep{DBLP:conf/tsd/AllisonGG06}. 

\begin{figure}[t]
     \centering
     \includegraphics[width=0.7\linewidth]{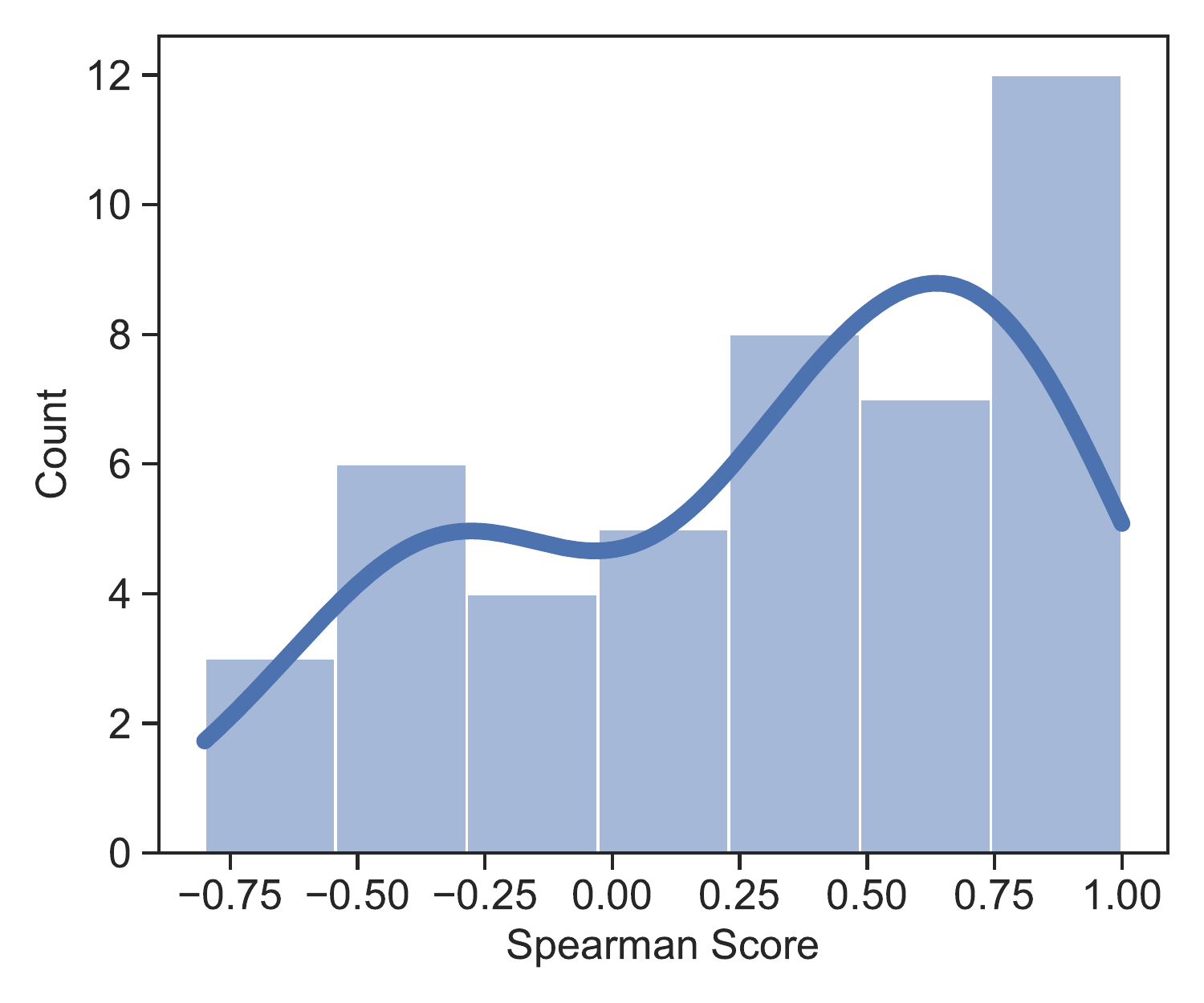}
     \caption{ MUV and downstream performance are positively correlated on two-thirds of tasks. X-axis classifies Spearman scores into different groups. Y-axis shows the number of tasks in each group. The middle Spearman score is 0.4. }
     \label{fig:mubleu}
 \end{figure}

To address this problem, we borrow the concept of \textit{Marginal Utility} in economics~\citep{samuelson1937note} and propose to use \textit{Marginal Utility of Vocabularization}~(MUV) as the optimization objective.
 MUV evaluates the benefits (entropy)  a corpus can get from an increase of cost (size). Higher MUV is expected for higher benefit-cost ratio.
Preliminary results verify that MUV correlates with downstream performances on two-thirds of translation tasks (See Figure~\ref{fig:mubleu}).
According to this feature, our goal turns to maximize MUV in tractable time complexity. 

\paragraph{Definition of MUV} Formally, MUV represents the negative derivation of entropy to size. For simplification, we leverage a smaller vocabulary to estimate MUV in implementation. Specially, MUV is calculated as:
\begin{equation}
\small
     \mathcal{M}_{v(k+m)} = \frac{-(\mathcal{H}_{v(k+m)}-\mathcal{H}_{v(k)})}{m},
    \label{eq:id}
\end{equation}
where  $v(k)$, $v(k+m)$ are two vocabularies with $k$ and $k+m$ tokens, respectively. 
$\mathcal{H}_v$ represents the corpus entropy with the vocabulary $v$, which is defined by the sum of token entropy. 
To avoid the effects of token length, here we normalize entropy with the average length of tokens and the final entropy is defined as:
\begin{equation}
\small
   \mathcal{H}_{v} = - \frac{1}{l_{v}}\sum_{j \in v } P(j)\log P(j), %\frac{1}{l_T}
\end{equation}
where $P(j)$ is the relative frequency of token $j$ from the training corpus and $l_{v}$ is the average length of tokens in vocabulary $v$.

\paragraph{Preliminary Results} To verify the effectiveness of MUV as the vocabulary measurement, we conduct experiments on 45 language pairs from TED and calculate the \textit{Spearman correlation score}\footnote{\url{https://www.statstutor.ac.uk/resources/uploaded/spearmans.pdf}} between MUV and BLEU scores. We adopt the same and widely-used settings to avoid the effects of other attributes on BLEU scores, such as model hyper-parameters and training hyper-parameters. We generate a sequence of vocabularies with incremental sizes via BPE.   All experiments use the same hyper-parameters. Two-thirds of pairs show positive correlations as shown in Figure~\ref{fig:mubleu}. The middle Spearman score is 0.4. We believe that it is a good signal to show MUV matters. Please refer to Section~\ref{sec:experiment} for more dataset details and Appendix A for more implementation details. 
%Experiment settings can be found in Section~\ref{sec:experiment}. 

%To be specific, 72.7% tasks show correlations where 60.0% tasks show over medorate positive correlations and 6.7% tasks show weak positive correlations. Considering that other factors (e.g. corpus size) also affect BLEU scores, we believe that it is a good evidence to support the claim.  Experiment settings can be found at Section~\ref{sec:experiment}. 

%We sample 3 language-pairs, and the results are listed in Figure~\ref{fig:find}. The full results on 12 language-pairs can be found in Appendix A. As we can see, vocabularies with the highest DeV  usually bring higher BLEU scores in most cases. Based on our full results, 42.8\% percent of language-pairs with the highest DeV achieve the best BLEU scores. Moreover, 35.7\% percent of language-pairs with the highest DeV almost achieve the best BLEU scores (gap less than 0.3). We calculate the Spearman correlation between two variables, DeV and BLEU.  $P > $ means two variables are strongly related in statistics. 

Given MUV, we have two natural choices to get the final vocabulary: search and learning. In the search-based direction, we can combine MUV with widely-used vocabularization solutions. For example, the optimal vocabularies can be obtained by enumerating all candidate vocabularies generated by BPE. While being simple and effective, it is not a self-sufficient approach. Furthermore, it still requires a lot of time to generate vocabularies and calculate MUV. To address these problems, we further explore a learning-based solution \method for more vocabulary possibilities. We empirically compare MUV-Search and \method in Section 5. 

%While being simple, the main limitation lies in the vast search space.  Assuming we have $N$ token candidates, the target of search-based approaches is to find the optimal vocabulary from $2^N$ subsets of tokens.  In a real-world scenario with limited resources,  we need a more efficient examination solution. Thus, we explore the learning-based direction to figure out ``how far an information-theoretic learning approach can reach''. 

% the number of explored vocabularies is limited due to the expensive cost to get vocabularies from merging operations, which depends on the scale of training data. By contrast, the learning direction has a higher performance ceiling.

%DeV evaluates  from information contribution. with the increase of vocabulary size, 

\section{Maximizing MUV via Optimal Transport}
\label{sec:problem}
This section describes the details of the proposed approach. We first show the general idea of \method in Section~\ref{sec:obj}, then describe the optimal transport solution in Section~\ref{sec:details}, followed by the implementation details in Section~\ref{sec:imp}. 

% and required to larger than the number of characters
\subsection{Overview}
\label{sec:obj}
We formulate vocabulary construction as a discrete optimization problem whose target is to find the vocabulary with the highest MUV according to Eq.~\ref{eq:id}. 
However, the vocabulary is discrete and such discrete search space is too large to traverse, which makes the discrete optimization intractable.

In this paper, we simplify the original discrete optimization problem by searching for the optimal vocabulary from vocabularies with fixed sizes.
Intuitively, MUV is the first derivative of entropy according to the vocabulary size (Eq.~\ref{eq:id}), and we introduce an auxiliary variable $\bm S$~($\bm S$ is an \textit{incremental} integer sequence) to approximate the computation by only computing MUV between vocabulary sizes as adjacent integers in $\bm S$.
% To tackle the discrete vocabulary, we estimate continuous MUV of vocabulary $v(k+m)$ with Eq. 1 based on the difference between the entropy of two vocabularies with size $k$, and $k+m$.  The whole search space is too big to be solved at acceptable costs. We simplify this question by searching for the optimal vocabulary from vocabularies with fixed sizes. %Since MUV is a dynamic feature depending on the amortized entropy difference between two vocabularies, we also formulate a dynamic process here.} \textcolor{red}{The dynamic process is used to search the vocabulary efficiently.

Formally, $\bm S = \{k, 2\cdot k, ..., (t-1)\cdot k, \cdots \}$ where each timestep $t$ represents a set of vocabularies with the number up to $\bm S[t]$. $k$ is the interval size. Of course, you can start from $\bm S[t] \geq |C|$ where $|C|$ is the size of all characters.  For any vocabulary, its MUV score can be calculated based on  a vocabulary from its previous timestep. We use $k$ to estimate the size gap between two vocabularies $v(t-1)$ and $v(t)$.  
With sequence $\bm S$, the target to find the optimal vocabulary $v(t)$ with the highest MUV can be formulated as:

{\small
\begin{align*}
&\argmax_{t}\argmax_{v(t-1)^ \in \mathbb{V}_{\bm S[t-1]}, v(t) \in \mathbb{V}_{\bm S[t]}} \mathcal{M}_{v(t)}  = \\
&\argmax_{t}\argmax_{v(t-1) \in \mathbb{V}_{\bm S[t-1]}, v(t) \in \mathbb{V}_{\bm S[t]}}  - \frac{1}{k} \big [ \mathcal{H}_{v(t)} - \mathcal{H}_{v(t-1)} \big ]
 \label{eq:reformulate}
\end{align*}}

% {\small
% \begin{align*}
%   & \argmax_{v(t-1)^ \in \mathbb{V}_{\bm S[t-1]}, v(t) \in \mathbb{V}_{\bm S[t]}} \mathcal{M}_{v(t)}  = \\  
%   &\argmax_{v(t-1) \in \mathbb{V}_{\bm S[t-1]}, v(t) \in \mathbb{V}_{\bm S[t]}}  - \frac{1}{k} \big [ \mathcal{H}_{v(t)} - \mathcal{H}_{v(t-1)} \big ],
%   \label{eq:reformulate}
% \end{align*}}

where $\mathbb{V}_{\bm S[t-1]}$ and $\mathbb{V}_{\bm S[t]}$  are two sets containing all vocabularies with upper bound of size $\bm S[t-1]$ and $\bm S[t]$.  The inner $\argmax$ represents that the target is to find the vocabulary from $\mathbb{V}_{\bm S[t]}$ with the maximum MUV scores. The outer $\argmax$ means that the target is to enumerate all timesteps and find the vocabulary with the maximum MUV scores. 

For a valid $\mathcal{M}$, the size of $v(t-1)$ is required to be smaller than $v(t)$. Actually, in our formulation the size of $v(t-1)$ may be larger than $v(t)$. Therefore, the search space contains illegal paris ($v(t-1)$, $v(t)$). Actually, following our IPC curves, we can find that smaller vocabulary usually have large $\mathcal{H}$. According to these findings, we can assume the following holds: 

{\small
\begin{align*}
& \argmax_{illegal (v(t-1), v(t))}-(\mathcal{H}_{v(t)} - \mathcal{H}_{v(t-1)}) \\
& \leq \argmax_{valid  (v(t-1), v(t))}-(\mathcal{H}_{v(t)} - \mathcal{H}_{v(t-1)}) 
\end{align*}}

Therefore, these illegal pair does not affect our results.  Due to exponential search space, we propose to optimize its  upper bound:
\begin{equation}
\small
    \argmax_{t}\frac{1}{k} \big [ \argmax_{v(t) \in \mathbb{V}_{\bm S[t]} }\mathcal{H}_{v(t)} -  \argmax_{v(t-1) \in \mathbb{V}_{\bm S[t-1]}} \mathcal{H}_{v(t-1)} \big ].
    \label{eq:solution}
\end{equation}
where $k$ means the size difference between $t-1$ vocabulary and $t$ vocabulary. 

\paragraph{Proofs} If we ignore the outer $\argmax$ and only consider the inner $\argmax$, we can get the following re-formulation:

{\small
\begin{align*}
\argmax_{v(t-1) \in \mathbb{V}_{\bm S[t-1]}, v(t) \in \mathbb{V}_{\bm S[t]}}  - \frac{1}{k} \big [ \mathcal{H}_{v(t)} - \mathcal{H}_{v(t-1)} \big ] & \Leftrightarrow   \\
 \argmin_{v(t-1) \in \mathbb{V}_{\bm S[t-1]}, v(t) \in \mathbb{V}_{\bm S[t]}} \big [ \mathcal{H}_{v(t)} - \mathcal{H}_{v(t-1)} \big ] & \Leftrightarrow    \\
 \argmin_{v(t) \in \mathbb{V}_{\bm S[t]}}\mathcal{H}_{v(t)} - \argmax_{v(t-1) \in \mathbb{V}_{\bm S[t-1]}}\mathcal{H}_{v(t-1)} &
\end{align*}}

% where the $k$ is a constant that does not affect the optimization, which can be ignored. We can further re-formulate the target as:

% {\small
% \begin{align*}
% &\argmin_{v(t-1) \in \mathbb{V}_{\bm S[t-1]}, v(t) \in \mathbb{V}_{\bm S[t]}} \big [ \mathcal{H}_{v(t)} - \mathcal{H}_{v(t-1)} \big ] \Leftrightarrow  \\
% & \argmin_{v(t) \in \mathbb{V}_{\bm S[t]}}\mathcal{H}_{v(t)} - \argmax_{v(t-1) \in \mathbb{V}_{\bm S[t-1]}}\mathcal{H}_{v(t-1)}
% \end{align*}}

and we can get the upper bound as shown in Eq.~\ref{eq:solution}.

Based on this equation, the whole solution is split into two steps: 1) searching for the optimal vocabulary with the highest entropy at each timestep $t$; 2) enumerating all timesteps and outputing the vocabulary corresponding to the time step satisfying Eq.~\ref{eq:solution}.  %Section~\ref{sec:details} show the details of the optimal transport solution in the first step. Section~\ref{sec:imp} shows the implementation details of \method. 
\

 \begin{figure}[t]
     \centering
     \includegraphics[width=\linewidth]{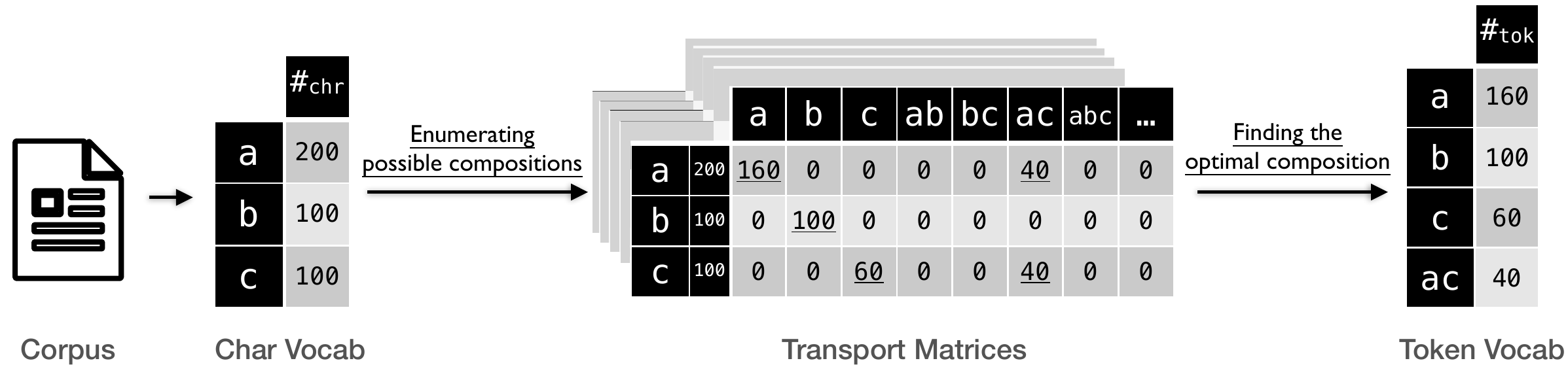}
     \caption{An illustration of vocabulary construction from a transport view. Each transport matrix represents a vocabulary. The transport matrix decides  how many chars are transported to token candidates. The tokens with zero chars will not be added into the vocabulary.  }
     \label{fig:ot}
 \end{figure}

%\subsection{Setup of \method}
%\label{sec:ot}
The first step of our approach is to search for the vocabulary with the highest entropy from  $\mathbb{V}_{\bm S[t]}$. Formally, the goal is to find a vocabulary $v(t)$ such that entropy is maximized, 
\begin{equation}
\small
    \argmax_{v(t) \in \mathbb{V}_{\bm S[t]} } -\frac{1}{l_{v(t)}}\sum_{j\in v(t)} P(j)\log P(j),
    \label{eq:step1}
\end{equation}
where 
$l_v$ is the average length for tokens in $v(t)$, $P(j)$ is the probability of  token $j$. However, notice that this problem is in general intractable due to the extensive vocabulary size. Therefore, we instead propose a relaxation in the formulation of discrete optimal transport, which can then be solved efficiently via the Sinkhorn algorithm~\citep{DBLP:conf/nips/Cuturi13}.

Intuitively, we can imagine vocabulary construction as a transport process that transports chars into token candidates with the number up to $\mathbf{S}[t]$.  As shown in Figure~\ref{fig:ot}, the number of chars is fixed, and not all token candidates can get enough chars. Each  transport matrix can build a vocabulary by collecting tokens with  chars. Different transport matrices bring different transport costs. The target of optimal transport is to find a transport matrix to minimize the transfer cost, i.e., negative entropy in our setting. %We implement an entropy-based Sinkhorn algorithm to solve this problem.

\subsection{Vocabularization via Optimal Transport}
\label{sec:details}
%%\paragraph{A tractable lower bound of entropy}
 Given a set of vocabularies $\mathbb{V}_{\bm S[t]}$, we want to find the vocabulary with the highest entropy. Consequently, the objective function in Eq.~\ref{eq:step1} becomes

{
\small
\begin{gather*}
\small
    \min_{v \in \mathbb{V}_{\bm S[t]} } \frac{1}{l_{v}}\sum_{j\in v} P(j)\log P(j), \\
\text{s.t.}\quad  P(j) = \frac{\text{Token}(j)}{\sum_{j\in v }\text{Token}(j)}, \,\, l_v = \frac{\sum_{j\in v} len(j)}{|v|}.
\end{gather*}}
 $\text{Token}(j)$ is the frequency of token $j$ in the vocabulary $v$. $len(j)$ represents the length of token $j$. 
Notice that both the distribution $P(j)$ and the average length $l_v$ depend on the choice of $v$.

% First, observe that since the final chosen token set will be large enough such that the statics $l_T$ is stable by the central limit theorem. In particular, we may assume there to be a uniform lower bound of $l_v$ for $v \in \mathbb{V}_{\bm S}$ when $S$ is sufficiently large. drop $l_T$ in the objective function without affecting the selection. 
\paragraph{Objective Approximation} To obtain a tractable lower bound of entropy, it suffices to give a tractable upper bound of the above objective function. We adopt the merging rules to segment raw text similar with BPE where two consecutive tokens will be merged into one if the merged one is in the vocabulary. 
To this end, let $\mathbb{T}\in \mathbb{V}_{\bm S[t]}$ be the vocabulary containing top $S[t]$ most frequent tokens, $\mathbb{C}$ be the set of chars and $|\mathbb{T}|, |\mathbb{C}|$ be their sizes respectively. 
Since $\mathbb{T}$ is an element of $\mathbb{V}_{\bm S[t]}$, clearly, we have

{
\small
\begin{align}\label{obj}	
	\min_{v \in \mathbb{V}_{\bm S[t]}} \frac{1}{l_v}\sum_{j\in v} P(j) \log P(j) & 
	 \leq \frac{1}{l_\mathbb{T}} \sum_{j \in \mathbb{T}} P(j) \log P(j).
\end{align}}
Here we start from the upper bound of the above objective function, that is $\frac{1}{l_\mathbb{T}} \sum_{j \in \mathbb{T}} P(j) \log P(j)$ and then search for a refined token set from $\mathbb{T}$. In this way, we reduce the search space into the subsets of $\mathbb{T}$. Let $P(j, i)$ be the joint probability distribution of the tokens and chars that we want to learn. Then we have
\begin{equation}
\small
    \begin{split}
	\sum_{j\in \mathbb{T}} P(j) \log P(j)& = \sum_{j\in \mathbb{T}} \sum_{i\in \mathbb{C}} P(j,i)\log P(j) \\
% 	& = \sum_{i\in \mathbb{T}} \sum_{j\in \mathbb{C}}P(i,j)\log P(i, j)\cdot \frac{P(i)}{P(i, j)} \\
% 	& = \sum_{i\in \mathbb{T}} \sum_{j\in \mathbb{C}}P(i,j)\log P(i, j) + \sum_{i\in \mathbb{T}} \sum_{j\in \mathbb{C}} P(i,j)\log \frac{P(i)}{P(i, j)} \\
	& = \underbrace{\sum_{j\in \mathbb{T}} \sum_{i\in \mathbb{C}} P(j,i)\log P(j, i)}_{\mathcal{L}_1}  \\
	& + \underbrace{ \sum_{j\in \mathbb{T}} \sum_{i\in \mathbb{C}}P(j,i)(-\log P(i | j))}_{\mathcal{L}_2}.\\
\end{split}
\label{eq:8}
\end{equation}
The details of proof can be found at Appendix C. Since $\mathcal{L}_1$ is nothing but the negative entropy of the joint probability distribution $P(j, i)$,  we shall denote it as $- H(P)$.

Let $\mD$ be the $|\mathbb{C}| \times |\mathbb{T}|$ matrix whose $(j, i)$-th entry is given by $-\log P(i | j)$, and let $\mP$ be the joint probability matrix, then
we can write
\begin{equation}
\small
    \mathcal{L}_2 = \left<\mP, \mD\right> = \sum_{j}\sum_{i}\mP(j,i)\mD(j,i).
\end{equation}

In this way, Eq.~\ref{eq:8} can be reformulated as the following objective function which has the same form as the objective function in optimal transport:
\begin{equation}\label{EOT}
\small
    \min_{\mP\in\mathbb{R}^{m\times n}} \left<\mP, \mD\right> - \gamma H(\mP).
\end{equation}

\paragraph{Setup of OT} From the perspective of optimal transport, $\mP$ can be regarded as the transport matrix, and $\mD$ can be regarded as the distance matrix.  Intuitively, optimal transport is about finding the best transporting mass from the char distribution to the target token distribution with the minimum work defined by $\left<\mP, \mD\right> $. 

To verify the validness of transport solutions, we add the following constraints. First, to avoid invalid transport between char $i$ and token $j$, we set the distance to  $+\infty$ if the target token $j$ does not contain the char $i$. Otherwise, we use $\frac{1}{len(j)}$ to estimate $ P(i | j)$ where $len(j)$ is the length of token $j$.
Formally, the distance matrix is defined as 
\[
\small
     \mD(j,i)= 
\begin{cases}
  -\log P(i | j) = +\infty,& \text{if }  i\notin j\\
   -\log P(i | j)  = -\log \frac{1}{len(j)},              & \text{otherwise}
\end{cases}
\label{eq:d}
\]
Furthermore, the number of chars is fixed and we set the sum of each row in the transport matrix to the probability of char $i$. The upper bound of the char requirements for each token is fixed and we set the sum of each column in the transport matrix to the probability of token $j$.
Formally, the constraints are defined as:
\begin{equation}
\small
   |\sum_i \mP(j, i) - P(j)| \leq \epsilon,
\end{equation}
and
\begin{equation}
\small
    \sum_{j}\mP(j,i) = P(i).
\end{equation}

% \begin{equation}
% \small
%     K_{ij} = -\log P(j | i) = +\infty, j\notin i
%     \label{eq:k}
% \end{equation}
%$K_{ij} = -\log P(j | i) = +\infty$ if $j\notin i$ and  $-\log \frac{\# c \,\in\, t}{len(t)}$ otherwise. 
% From the perspective of optimal transport, $\mP$ is the transport matrix and $\mK$ is the distance matrix.  $+\infty$ is introduced to ensure that the transport matrix only distributes valid chars to token candidates.  Also, for transport matrix, we have the hard constraints $\sum_{j} P(i, j) = P(i)$ and $\sum_i P(i, j) = P(j)$ where $P(i), P(j)$ are the char distribution and candidate token distribution of $\mathbb{T}$, respectively.
%However, in order to obtain a refined token set from $\mathbb{T}$ with larger entropy, we need to relax the hard constraint on the token distribution matching to a soft constraint. This formulation then allows us to drop out tokens with low joint probability distribution. 

\begin{figure}[t]
     \centering
     \includegraphics[width=0.9\linewidth]{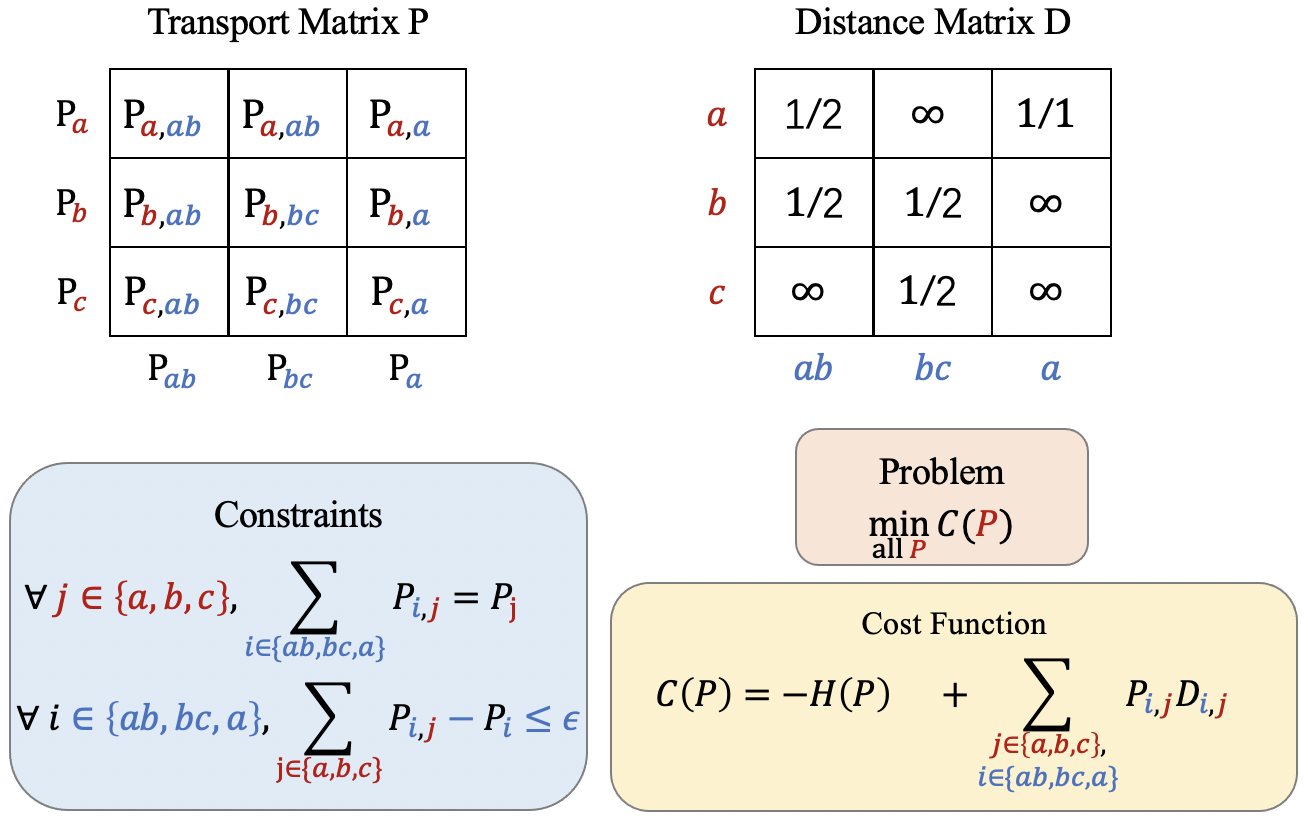}
     \caption{The details of optimal transport. The objective function is the sum of negative entropy and transport cost. Each element $D(i,j)$ in the distance matrix is the negative log of $1/n$ where $n$ is the length of token $i$. It defines the distance between char $j$ and token $i$.  To avoid invalid transport between char $j$ and token $i$, we set the distance to infinite if the target token $i$ does not contain the char $j$. } 
     \label{fig:otdetail}
 \end{figure}

Given transport matrix $\mP$ and distance matrix $\mD$, the final objective can be formulated as:

{
\small
   \begin{gather*}
        \argmin_{\mP \in \R^{|\mathbb{C}| \times |\mathbb{T}|}} - H(\mP) + \left<\mP, \mD\right>, \\
 \text{s.t.}\quad  \sum_{i} \mP(i, j) = P(j), \quad |\sum_j \mP(i, j) - P(i)| \leq \epsilon,
   \end{gather*}}
   %, \quad \forall i, j
 with small $\epsilon > 0$. Figure~\ref{fig:otdetail} shows the details of optimal transport solution.   Strictly speaking, this is an unbalanced entropy regularized optimal transport problem. Nonetheless, we can still use the generalized Sinkhorn algorithm to efficiently find the target vocabulary as detailed in Section 4.6 of ~\citet{peyre2020computational}. The algorithm details are shown in Algorithm~\ref{alg:ot}. At each timestep $t$, we can generate a new vocabulary associated with entropy scores based on the transport matrix $\mP$. Finally, we collect these vocabularies associated with entropy scores, and output the vocabulary satisfying Eq.~\ref{eq:solution}.

\begin{algorithm}[t]
\footnotesize
%\SetAlgoLined
\caption{\method}
\textbf{Input:} A sequence of token candidates $\mathbb{L}$ ranked by frequencies, an incremental integer sequence $\bm S$ where the last item of $\bm S$ is less than $|\mathbb{L}|$, a character sequence $\mathbb{C}$, a training corpus $D_{c}$  \\
\textbf{Parameters:} \textrm{u} $\in \R_+^{|\mathbb{C}|}$, \textrm{v} $\in \R_+^{|\mathbb{T}|}$   \\
vocabularies = [] \\
\For{ item in  $\bm S$}{ \tcp{Begin of Sinkhorn algorithm }
Initialize \textrm{u} = ones() and \textrm{v} = ones() \\

$\mathbb{T} = \mathbb{L}[:item]$  \\
 Calculate token frequencies $P(\mathbb{T})$ based on $D_{c}$ \\
Calculate char frequencies $P(\mathbb{C})$ based on $D_{c}$ \\
Calculate $\mD$\\
\While{not converge}{
\textrm{u} = $ P(\mathbb{T}) / \mD\textrm{v} $\\
\textrm{v} = $P(\mathbb{C}) / \mD^T \textrm{u} $
}
optimal\_matrix = \textrm{u}.reshape(-1, 1) * $\mD$ * \textrm{v}.reshape(1, -1)
\\
 \tcp{End of Sinkhorn algorithm }
 entropy, vocab = get\_vocab(optimal\_matrix) \\
 %\tcp{Generate a vocabulary based on the transport matrix}
vocabularies.append(entropy,vocab) \\ 
}
Output $\bm v^*$ from vocabularies satisfying Eq.~\ref{eq:solution}

%\textbf{Output:$\bm v^*$ }
\label{alg:ot}
\end{algorithm}

\subsection{Implementation}
\label{sec:imp}
Algorithm~\ref{alg:ot} lists the process of \method. First, we rank all token candidates according to their frequencies. For simplification, we adopt BPE-generated tokens  (e.g. BPE-100K) as the token candidates. It is important to note that any segmentation algorithms can be used to initialize token candidates. Experiments show that different initialization approaches result in similar results. We simply adopt BPE-100K for bilingual translation and BPE-300K for multilingual translation in this work. All token candidates with their probabilities are then used to initialize $\mathbb{L}$ in Algorithm~\ref{alg:ot}.   The size of the incremental integer sequence $\bm S$ is a hyper-parameter and set to $(1K, ..., 10K)$ for bilingual translation, $(40K, ..., 160K)$ for multilingual settings. %For $\mathbb{V}_{\bm S[t]}$, we use  $\bm S_{bpe}[:\bm S[t]]$ as the target token candidate distribution. For char distribution, we use char probability in the raw text.
%Due to the large space of token candidates, we adopt BPE generated tokens  (e.g. BPE-100K) as the target token candidates. It is important to note that any segmentation algorithms can be used to initialize token candidates. Experiments show that different initialization results in similar results and we simply adopt BPE-100K in this work.  Each merge action represents a new token. We range all tokens based on the generation rank. In this way, we can get a sequence of tokens associated with their
At each timestep, we can get the vocabulary with the maximum entropy based on the transport matrix.  It is inevitable to handle illegal transport case due to relaxed constraints. We remove tokens with distributed chars less than $0.001$ token frequencies. Finally, we enumerate all timesteps and select the vocabulary satisfying Eq.~\ref{eq:solution} as the final vocabulary. 

After generating the vocabulary,  \method uses a greedy strategy to encode text similar to BPE. To encode text, it first splits sentences into character-level tokens. Then, we merge two consecutive tokens into one token if the merged one is in the vocabulary. This process keeps running until no tokens can be merged. Out-of-vocabulary tokens will be split into smaller tokens. 

%Then, based on BPC values in different timesteps,  the optimal DeV score can be obtained via Eq.~\ref{eq:solution}. Finally, according to the transport matrix at the timestep with the highest DeV, we generate the final vocabulary via a post-processing pipeline. 

%\paragraph{Implementation Details}
% Given takes these two distributions as input and generate a vocabulary 

%\section{Info-VOT: The \method Approach}
%\label{sec:approach}
%\input{045approach.tex}

\section{Experiments}
\label{sec:experiment}

To evaluate the performance of \method, we conduct experiments on three datasets, including WMT-14 English-German translation,  TED bilingual translation, and TED multilingual translation.  

\subsection{Settings}

We run experiments on the following machine translation datasets. See Appendix B for more model and training details. %$Appendix 1 shows an detailed description of these datasets.

\begin{enumerate}
    \item WMT-14 English-German (En-De) dataset: This dataset has 4.5M sentence pairs. The dataset is processed following~\citet{scalingnmt}. 
We choose newstest14 as the test set. 
    
    % \item WMT-16 English-Romanian (En-Ro) dataset: It has 2.2M training sentence pairs. We use the officially released validation and test sets.  We adopt the same pre-processing pipeline in the WMT14 En-De dataset.
    
    \item TED bilingual dataset: We include two settings: X-to-English translation and English-to-X translation. We choose 12 language-pairs with the most training data. We use the language code according to ISO-639-1 standard\footnote{http://www.lingoes.net/en/translator/langcode.htm}. TED data is provided by~\citet{Ye2018WordEmbeddings}. %We run experiments with 50 epochs and average the last 5 checkpoints as the final network for evaluation.
    
    \item TED multilingual dataset: We conduct experiments with 52 language pairs on a many-to-English setting. The network is trained on all language pairs. We adopt the same pre-processing pipeline in the WMT-14 En-De dataset. 
\end{enumerate}

\subsection{Main Results}

%\subsection{Results}

%#######################################################################################################################################
% BPE-30K is the most popular setting in 42 papers accepted by the research track of Conference of Machine Translation (WMT) through 2017 and 2018.  * means WMT translation results. The rest columns are TED results. 
\begin{table*}[ht]
    \centering
    \footnotesize
    \setlength{\tabcolsep}{3.2pt}
    \caption{Comparison between vocabularies search by \method and widely-used BPE vocabularies. \method achieves higher BLEU scores with large size reduction. Here the vocabulary size is adopted from the X-En setting. }
    \begin{tabular}{l|c|cccccccccccc}
    \toprule
    Bilingual & \multicolumn{1}{c|}{ WMT-14} & \multicolumn{12}{c}{ TED} \\
    \midrule
    \multirow{1}{*}{ En-X}   &  De &  Es &  PTbr &  Fr &  Ru & He &  Ar &    It &  Nl &  Ro &  Tr  &  De & Vi  \\ \midrule 
       
        BPE-30K & 29.31 & 39.57 & 39.95 & 40.11 & 19.79  & 26.52 &16.27 & 34.61 & 32.48 & 27.65 & 15.15 & 29.37 & 28.20 \\
        \bf \method & \bf 29.80  & \bf 39.97 & \bf 40.47 & \bf 40.42 &  \bf 20.36 & \bf 27.98 &  \bf 16.96    & \bf 34.64  &  \bf 32.59  &  \bf  28.08  & \bf  16.17 & \bf 29.98 &  \bf 28.52 \\
    \toprule
      %\multirow{2}{*}{\bf X-En } & \multicolumn{1}{c|}{\bf WMT-14} & \multicolumn{12}{c}{\bf TED} \\ 
      X-En  &  De &  Es &  PTbr & Fr & Ru &  He &  Ar  & It &  Nl &  Ro &  Tr  &  De & Vi \\ \midrule 
      
        BPE-30K & \bf 32.60 & \bf 42.59 & 45.12 & \bf  40.72 & 24.95 & 37.49 & 31.45 & 38.79 & 37.01 & 35.60 & 25.70 & 36.36 & 27.48 \\
       \bf \method & 32.30     &  42.34 & \bf 45.93  & \bf 40.72 & \bf 25.33 &  \bf 38.70 & \bf 32.97 & \bf 39.09 & \bf 37.31 & \bf 36.53 & \bf 26.75 & \bf 36.68 & \bf 27.39 \\
        \toprule
        % \multirow{2}{*}{\bf Vocab Size (K)} & \multicolumn{1}{c|}{\bf WMT-14} & \multicolumn{12}{c}{\bf TED} \\ 
    Vocab Size (K) &  De &  Es &  PTbr &  Fr &  Ru &  He &  Ar  &  It &  Nl &  Ro &  Tr  &  De & Vi \\ \midrule 
        
        BPE-30K & 33.6  & 29.9 & 29.8 & 29.8 & 30.1 & 30.0 & 30.3 & 33.5 & 29.8 & 29.8 & 29.9 & 30.0 & 29.9  \\
       \bf \method &  \bf 11.6 & \bf  5.3 & \bf 5.2 & \bf 9.2 & \bf 3.3 & \bf 7.3 & \bf 9.4 &\bf 3.2 & \bf 2.4 & \bf 3.2 & \bf 7.2 &  \bf 8.2 & \bf  8.4 \\
   
       \bottomrule

    \end{tabular}

    \label{tab:bpe}
\end{table*}

%###########################

%#######################################################################################################################################
\begin{table*}[t]
    \centering
    \footnotesize
    \setlength{\tabcolsep}{3.2pt}
    \caption{Comparison between vocabularies search by \method and BPE-1K, recommended by~\citet{DBLP:conf/mtsummit/DingRD19} for low-resource datasets. Here we take TED X-En bilingual translation as an example. This table demonstrates that vocabularies searched by \method are on par with heuristically-searched vocabularies in terms of BLEU scores. }
   
    \begin{tabular}{l|cccccccccccc|c}
    \toprule
   
     X-En  & Es &  PTbr &  Fr & Ru &  He &  Ar &  It & Nl &  Ro &  Tr  & De & Vi &  Avg \\ \midrule 
      
        BPE-1K  & \bf 42.36 & 45.58 & \bf 40.90 & 24.94 &38.62 &  32.23 & 38.75 & \bf 37.44 & 35.74 & 25.94 & \bf 37.00 & 27.28  & 35.65 \\
      \bf \method &  42.34 & \bf 45.93  & 40.72 & \bf 25.33 &  \bf 38.70 & \bf 32.97 & \bf 39.09 & 37.31 & \bf 36.53 & \bf 26.75 &  36.68 & \bf 27.39 & \bf 35.81 \\
       \toprule
         Vocab Size (K) & Es & PTbr & Fr & Ru &  He &  Ar &  Ko & It &  Nl &  Ro &  Tr  &  De &  Avg  \\ \midrule 
        
       BPE-1K & \bf 1.4 & \bf 1.3 & \bf 1.3 & \bf 1.4 & \bf 1.3 & \bf 1.5 & \bf 4.7 & \bf 1.2 & \bf 1.2 & \bf 1.2 & \bf 1.2 & \bf 1.2 & \bf 1.6 \\
        \bf \method & 5.3 &5.2 &  9.2 &  3.3 & 7.3 & 9.4 & 3.2 & 2.4 & 3.2 &  7.2 &  8.2 &   8.4 & 6.0 \\
      %\bf \method &   2.4 & 2.2 &  2.1 &  2.3 &  2.2 &  2.4 & 5.6 &  2.1 &  1.9 & 2.1 &   2.1 &  1.9 & 2.6 \\

       \bottomrule

    \end{tabular}
%\vspace{-0.2cm}
    \label{tab:bpe1k}
\end{table*}

%###########################

\paragraph{ Vocabularies Searched by \method are Better than Widely-used Vocabularies on Bilingual MT Settings.}  

%To be specific, \method brings significant performance gains with over 80\%  vocabulary size reduction on TED.

~\citet{DBLP:conf/mtsummit/DingRD19} gather 42 papers that have been accepted by the research track of Conference of Machine Translation (WMT) through 2017 and
2018. Among these papers, the authors find that 30K-40K is the most popular range for the number of BPE merge actions. Following this work, we first compare our methods with dominant BPE-30K.  The results are listed in Table~\ref{tab:bpe}. As we can see, the vocabularies searched by \method achieve higher BLEU scores with large size reduction. 
 The promising results demonstrate that \method is a practical approach that can find a well-performing vocabulary with higher BLEU and smaller size.% We also conduct experiments on multilingual translation. Table 7 in Appendix B shows that \method beats widely-used vocabularies on 27 out of 39 datasets. 

\paragraph{ Vocabularies Searched by \method are on Par with Heuristically-searched Vocabularies on Low-resource Datasets.}  ~\citet{DBLP:conf/mtsummit/DingRD19} study how the size of BPE affects the model performance in low-resource settings. They conduct experiments on four language pairs and find that smaller vocabularies are more suitable for low-resource datasets. For Transformer architectures, the
optimal vocabulary size is less than 4K, around up to 2K merge actions. We compare \method and BPE-1K on an X-to-English bilingual setting. The results are shown in Table~\ref{tab:bpe1k}. We can see that \method can find a good vocabulary on par with heuristically searched vocabularies in terms of BLEU scores. Note that BPE-1K is selected based on plenty of experiments. In contrast, \method only requires one trials for evaluation and only takes  0.5 CPU hours plus 30 GPU hours to find the optimal vocabulary.

%#######################################################################################################################################
\begin{table*}[ht]
    \centering
    \footnotesize
    \setlength{\tabcolsep}{5pt}
    \caption{Comparison between \method and widely-used BPE vocabularies on multilingual translation.  \method achieves higher BLEU scores on most pairs.  }
    \begin{tabular}{l|ccccccccccccc}
    \toprule
      X-En &  Es &  Pt-br &  Fr &  Ru & He &  Ar &  Ko &  Zh-cn &  It & Ja & Zh-tw &  Nl & Ro   \\ 
     \midrule
     BPE-60K & 32.77 & 35.97 & 31.45 & 19.39 & 26.65 & \bf 22.28 & 14.13 & 15.80 & 29.06 & 10.31 & 15.03 & 26.83 & 26.44 \\
     \bf \method & \bf 33.84 & \bf 37.18 & \bf 32.85 & \bf 20.23 & \bf 26.85 & 22.17 & \bf 14.36 & \bf 16.59 & \bf 30.44 & \bf 10.75 & \bf 15.73 & \bf 27.68 & \bf 27.45 \\
     \toprule

      X-En & Tr  & De &  Vi &  Pl & Pt  & Bg  & El  & Fa  &  Sr  &  Hu &  Hr &  Uk &  Cs  \\
      \midrule
     BPE-60K & 16.74 & 25.92 & 21.00 &  18.06 & 34.17 & 30.41 & 29.35 & 20.49 & 26.66 & 17.97 & 28.30 & 22.18 & 22.08 \\
     \bf \method & \bf 17.55 & \bf 27.01  & \bf 22.25  & \bf 18.93  & \bf 35.64  & \bf 31.77  & \bf 31.27 &  \bf 20.05  & \bf 27.45  & \bf 19.00 & \bf 29.25  & \bf 23.34  & \bf 23.54 \\
     \toprule

    X-En &  Id   &  Th &  Sv &  Sk & Sq &  Lt &  Da &  My &  Sl &  Mk &  Fr-ca &  Fi & Hy   \\
   \midrule
    BPE-60K &24.58 &  17.92 & 30.43 & 24.68 & 28.50 & 19.17 & 34.65 & 13.54 & 20.59 & 28.23 & 27.20 & 15.13 & 17.68 \\
    \bf \method  &  \bf 25.87 & \bf 18.89  &  \bf 31.47  & \bf 25.69 & \bf 29.09  & \bf 19.85  & \bf 36.04 & \bf 13.65 & \bf 21.36  &\bf 28.54  &\bf 28.35 & \bf 15.98  & \bf 18.44 \\
    
     \toprule
       X-En & Hi & Nb & Ka & Mn & Et & Ku & Gl & Mr & Zh & Ur & Eo & Ms & Az\\
       \midrule
        BPE-60K & \bf 18.57 & \bf 35.96& \bf 16.47& \bf 7.96 & 15.91 & \bf 13.39 & 26.75 & \bf 8.94 & \bf 13.35 & \bf 14.21 &\bf  21.66 & \bf 19.82 & \bf 9.67 \\
         \bf \method  & 18.54 & 35.88 & 15.97 & \bf 7.96 & \bf 16.03 & 13.20 &\bf 26.94 & 8.40 & 12.67 & 13.89 & 21.43 & 19.06 & 9.09  \\

       \bottomrule

    \end{tabular}

    \label{tab:mul}
\end{table*}

%###########################

%, but still   takes a lot of time to generate vocabularies due to a large training set
\begin{table}[t]
\caption{Results of \method, MUV-Search and BPE-Search. MUV-Search does not require full training and saves a lot of costs.  Among them, \method is the most efficient solution. MUV-Search  and \method require additional costs for downstream evaluation, which takes around 32 GPU hours. ``GH'' and ``CH'' represent GPU hours and CPU hours, respectively. }
\label{DeV-table}
\begin{center}
\footnotesize
\begin{tabular}{lccc}
\toprule
 En-De &  BLEU & Size &   Cost \\

\midrule
BPE-Search & \bf 29.9 & 12.6K & 384 GH \\
MUV-Search & 29.7 & \bf 9.70K &  5.4 CH + 30 GH\\
%VOT w.o Transformer (Base) &  & 8.5K & 8.7K &   \\
\bf \method &  29.8 &  11.6K &  \bf 0.5 CH + 30 GH\\% \bf 8.7K &

%VOT-Conv &  & 8.5K & 8.7K &  \\
\bottomrule
\end{tabular}
\end{center}
\label{tab:DeV}
\end{table}

%##########################################################################################################

\begin{table}[t]
\caption{Comparison between \method and strong baselines. \method achieves almost the best performance with a much smaller vocabulary.  }
\label{sample-table}
\begin{center}
\footnotesize
\begin{tabular}{lcc}
\toprule
 En-De &  BLEU  &  Parameters \\

\midrule

\cite{attention} & 28.4   & 210M \\ %& 33.6K
\cite{DBLP:conf/naacl/ShawUV18} &  29.2    & 213M \\% & 33.6K
\cite{scalingnmt} & 29.3   & 210M  \\ % 33.6K 
\cite{DBLP:conf/icml/SoLL19} & 29.8   & 218M \\ %& 33.6K
\cite{deeptransformer} & \bf 30.1  &  256M \\%33.6K &
%\midrule

%BPE w.o. Conv  & & 32K & 33.6K & \\
\midrule
%BPE w.o Transformer (Base) & & 32K & 33.6K &  \\
SentencePiece &  28.7   & 210M \\%& 33.6K
WordPiece & 29.0   & 210M \\%& 33.6K
%DeV-Search & 29.7 & 6K & 9K & 186M \\
%BPE-30K & 29.3 & 32K & 33.6K & 210M \\
\midrule
%VOT w.o Transformer (Base) &  & 8.5K & 8.7K &   \\
\bf \method &  29.8  & \bf 188M \\% \bf 8.7K &

%VOT-Conv &  & 8.5K & 8.7K &  \\
\bottomrule
\end{tabular}
\end{center}
\label{tab:sota}
\end{table}

\begin{table}[ht]
\caption{ Vocabularies searched by \method are better than widely-used vocabularies on various architectures. Here ``better'' means competitive results but much smaller sizes. }
\begin{center}
\footnotesize
\begin{tabular}{lccc}
\toprule
En-De & Approach &  BLEU &   Size  \\

\midrule

%BPE w.o Transformer (Base) & & 32K & 33.6K &  \\
\multirow{2}{*}{Transformer-big} & BPE-30K  & 29.3 & 33.6K  \\
& \bf \method &\bf 29.8 & \bf 11.6K  \\
%\midrule
% \multirow{2}{*}{Transformer} & BPE-30K & \bf 27.7 & 33.6K \\
% & \bf \method& 27.5 & \bf 8.0K  \\
\midrule
\multirow{2}{*}{Convolutional Seq2Seq} & BPE-30K &  \bf 26.4 & 33.6K \\
& \bf  \method & 26.3 & \bf 11.6K \\

%VOT-Conv &  & 8.5K & 8.7K &  \\
\bottomrule
\end{tabular}
\end{center}
%\vspace{-0.5cm}
\label{tab:ablation}
\end{table}

%Unlike BPE-Search, it does not require expensive GPU for full training. 

 \paragraph{\method Works Well on Multilingual MT Settings.}
 We conduct a multilingual experiment.  These languages come from multiple language families and have diverse characters.  We compare \method with BPE-60K, the most popular setting in multilingual translation tasks. Table~\ref{tab:mul} lists the full results.  The size of the searched vocabulary is around 110K. As we can see,  \method achieves better BLEU scores on most pairs.

\paragraph{\method is a Green Vocabularization Solution. } One advantage of \method lies in its low resource consumption. We compare \method with BPE-Search, a method to select the best one from a BPE-generated vocabulary set based on their BLEU scores. The results are shown in Table~\ref{tab:DeV}. In BPE-Search, we first define a vocabulary set including BPE-1K, BPE-2K, BPE-3K, BPE-4K, BPE-5K, BPE-6K, BPE-7K, BPE-8K, BPE-9K, BPE-10K, BPE-20K, BPE-30K. Then, we run full experiments to select the best vocabulary. Table~\ref{tab:DeV} demonstrates that \method is a lightweight solution that can find a competitive vocabulary within 0.5 hours on a single CPU, compared to BPE-Search that takes hundreds of GPU hours. The cost of BPE-Search is the sum of the training time on all vocabularies.  Furthermore, we also compare \method with MUV-Search as introduced in Section 3. MUV-Search is a method that combines MUV and popular approaches by selecting the vocabulary with the highest MUV as the final vocabulary. We generate a sequence of BPE vocabularies with incremental size 1K, 2K, 3K, 4K, 5K, 6K, 7K, 8K, 9K, 10K, 20K. For $t$-th vocabulary $v(t)$, its MUV score is calculated according to $v(t)$ and $v(t-1)$. We enumerate all vocabularies and select the vocabulary with the highest MUV as the final vocabulary. The comparison between \method and MUV-Search is shown in Table~\ref{tab:DeV}. Although MUV-Search does not require downstream full-training, it still takes a lot of time to generate vocabularies and calculate MUV. Among them, VOLT is the most efficient approach.

\subsection{Discussion}
We conduct more experiments to answer the following questions: 1) can a baseline beat strong approaches with a better vocabulary; 2) can \method beat recent vocabulary solutions, like SentencePiece; 3) can \method work on diverse architectures?

\paragraph{A Simple Baseline with a \method -generated Vocabulary Reaches SOTA Results.}
We compare \method and several strong approaches on the En-De dataset. Table~\ref{tab:sota} shows surprisingly good results. Compared to the approaches in the top block, \method achieves almost the best performance with a much smaller vocabulary. These results demonstrate that a simple baseline can achieve good results with a well-defined vocabulary. %In future work, we may seriously consider the best practice of benchmark approaches. 

\paragraph{\method Beats SentencePiece and WordPiece.} SentencePiece and WordPiece are two variants of sub-word vocabularies. We also compare our approach with them on WMT-14 En-De translation to evaluate the effectiveness of \method. The middle block of Table~\ref{tab:sota} lists the results of SentenPiece and WordPiece. We implement these two approaches with the default settings. We can observe that \method  outperforms SentencePiece and WordPiece by a large margin, with over 1 BLEU improvements.

\paragraph{\method Works on Various Architectures.} This work mainly uses Transformer-big in experiments. We are curious about whether \method  works on other architectures. We take WMT-14 En-De translation as an example and implement a Convolutional Seq2Seq model.  The network uses the default settings from Fairseq\footnote{\url{https://github.com/pytorch/fairseq/tree/master/examples/translation}}. We set the maximum epochs to 100 and average the last five models as the final network for evaluation. Table~\ref{tab:ablation} demonstrates that vocabularies searched by \method also works on Convolutional Seq2Seq  with competitive BLEU but much smaller size. In this work, we verify the effectiveness of \method on architectures with standard sizes. Since model capacity is also an important factor on BLEU scores, we recommend larger vocabularies associated with more embedding parameters for small architectures. 

\paragraph{\method can Bring Slight Speedup During Training.} We evaluate the running time for VOLT vocabulary and BPE-30K on WMT En-De translation. The model with VOLT-searched vocabulary (11.6k tokens) can process 133 sentences per second, while the model with BPE-30K (33.6k tokens) only executes 101 sentences per second. All experiments run on the same environment (2 Tesla-V100-GPUs + 1 Gold-6130-CPU), with the same beam size for decoding.  
The speedup mainly comes from larger batch size with reduced embedding parameters. We also find that although VOLT reduces the Softmax computations,  it does not significantly boost the Softmax running time due to optimized parallel computation in GPUs.   %Since VOLT mainly reduces the overhead of vocabulary searching, it is accepa significantly boost the training and inference speed due to optimized parallel computation in GPU.

\paragraph{\method Vocabularies and BPE Vocabularies are Highly Overlapped.} For simplification, \method starts from BPE-segmented tokens. We take WMT En-De as an example to see the difference between \method vocabulary and BPE vocabulary. The size of \method vocabulary is around 9K and we adopt BPE-9K vocabulary for comparison. We find that these two vocabularies are highly overlapped, especially for those high-frequency words. They also have similar downstream performance. Therefore, from an empirical perspective, BPE with \method size is also a good choice.

\section{Conclusion}
\label{sec:conclusion}
In this work, we propose a new vocabulary search approach without trail training. The whole framework starts from an informtaion-therotic understanding. According to this understanding, we formulate vocabularization as a two-step discrete optimization objective and propose a principled optimal transport solution \method. Experiments show that \method can effectively find a well-performing vocabulary in diverse settings. 

\section*{Acknowledgments}
We thank the anonymous reviewers, Demi Guo, for their
helpful feedback. Lei Li and Hao Zhou are corresponding authors.

\bibliographystyle{acl_natbib}
\bibliography{refs}

\clearpage

\section*{Appendix A: MUV}
To evaluate the relationship between MUV and BLEU scores, we conduct experiments on 45 language pairs (X-En) with most resources (including ar-en, eg-en, cs-en, da-en, de-en, el-en, es-en, et-en, fa-en, fi-en, fr-ca-en, fr-en, gl-en, he-en, hi-en, hr-en, hu-en, hy-en, id-en, it-en, ja-en, ka-en, ko-en, ku-en, lt-en, mk-en, my-en, nb-en, nl-en, pl-en, pt-br-en, pt-en, ro-en, ru-en, sk-en, sl-en, sq-en, sr-en, sv-en, th-en, tr-en, uk-en, vi-en, zh-cn-en, zh-tw-en) from TED and calculate the Spearman correlation score beween MUV and BLEU. We merge all bilingual training data together and pre-train a multilingual network. 
To avoid the effects of unsteady BLEU scores,  we use the multilingual network to initialize bilingual networks. All bilingual datasets are segment by four multilingual vocabularies, including BPE-20K, BPE-60K, BPE-100K, BPE-140K. In this way, we can get four bilingual corpora for each translation task. The MUV is calculated based on these corpora. For each corpus, we leverage a corpus with a smaller vocabulary to calculate MUV. For example, the MUV score of Ar-En (BPE-20K) is calculated based on Ar-En (BPE-20K) and Ar-En (BPE-10K).  It is important to note that all corpora adopt the same interval, 10K, to calculate MUV.  All bilingual datasets share the same model hyper-parameters and training hyper-parameters (Please refer to Appendix B for more implementation details). We set the maximum training epoch to 50 and average the last five models as the final network for evaluation. 
\section*{Appendix B: Experiments}

\paragraph{Models.} We use Fairseq to train a Transformer-big model with the same setting in the original paper~\citep{scalingnmt}. The input embedding and output embeddings are shared. We use the Adam optimizer~\citep{DBLP:journals/corr/KingmaB14} with a  learning rate 5e-4 and an inverse\_sqrt decay schedule. The warm-up step is $4,000$,  the dropout rate is $0.3$, the update frequency is $4$, the number of tokens is $9,600$, or $4,800$ in a single batch.

\paragraph{Training and Evaluation.} We run WMT-14 En-De experiments with 8 GPUs, TED bilingual translation with 4 GPUs, TED multilingual translation with 16 GPUs. We set a beamwidth to 4 for En-De and 5 for the other. For bilingual translation,  we run approaches 40 epochs, average the last five models on all datasets, and use the averaged model to generate translation results. For multilingual translation, all approaches run 10 epochs and we adopt the last model for evaluation. We calculate case-sensitive tokenized BLEU for evaluation.

% \paragraph{\method Works on Various Architectures.}  We take WMT'14 En-De as an example and implement a Transformer network and a Convolutional Seq2Seq network.  All networks use the default settings from Fairseq\footnote{https://github.com/pytorch/fairseq/tree/master/examples/translation}. We set the maximum epochs to 100 and average the last five models as the final network for evaluation. Table~\ref{tab:ablation} shows that  \method can find better vocabularies than widely-used vocabularies on diverse architectures. 

% \paragraph{\method can Find Better Vocabularies on Multilingual Translation. }
% Table~\ref{tab:mul} lists the comparison on multilingual translation.  These languages come from multiple language families and have diverse characters.   BPE-60K is the most popular setting in multilingual translation tasks. As we can see,  \method achieves better BLEU scores on 30 out of 45 language pairs.

\clearpage
\section*{Appendix C: Proofs for Eq. 6}

\begin{equation*}
    \begin{split}
	\sum_{i\in \mathbb{T}} P(j) \log P(j)& = \sum_{j\in \mathbb{T}} \sum_{i\in \mathbb{C}} P(j,i)\log P(j) \\
	& = \sum_{j\in \mathbb{T}} \sum_{i\in \mathbb{C}}P(j,i)\log P(j, i)\cdot \frac{P(j)}{P(j, i)} \\
	& = \sum_{j\in \mathbb{T}} \sum_{i\in \mathbb{C}}P(j,i)\log P(j, i) + \sum_{j\in \mathbb{T}} \sum_{i\in \mathbb{C}} P(j,i)\log \frac{P(j)}{P(j, i)} \\
	& = \underbrace{\sum_{j\in \mathbb{T}} \sum_{i\in \mathbb{C}} P(j,i)\log P(j, i)}_{\mathcal{L}_1} + \underbrace{ \sum_{j\in \mathbb{T}} \sum_{i\in \mathbb{C}}P(j,i)(-\log P(i | j))}_{\mathcal{L}_2}.\\
\end{split}
\end{equation*}

\qedsymbol

% \clearpage

% \section*{Appendix A: MUV}
% \input{070appendix.tex}
% \clearpage

% \section*{Appendix B: Experiments}
% \input{080appendix1.tex}

% \clearpage
% \section*{Appendix C: Proofs for Eq. 6}
% \input{090appendix2.tex}

%\clearpage
%\section*{Appendix D: More examples about Pareto optimality}
%\input{092appendix4.tex}

% \clearpage
% \section*{Appendix D: Supplemental Experiments}
% \input{091appendix3.tex}

\end{document}